\documentclass[11pt]{article}
\usepackage{xspace}
\usepackage{cprotect}
\usepackage{amsmath}
\usepackage{natbib}
\usepackage{graphicx}
\usepackage{tikz-qtree}

\newcommand{\alternativeMergeOrder}{\texttt{AlternativeMergeOrder}\xspace}
\newcommand{\lt}{<}

\title{A Statistical Comparison of Some Theories of NP~Word~Order}
\author{Richard Futrell, Roger Levy, and Matthew Dryer}
\date{\today}

\begin{document}
\maketitle
 
\begin{abstract}
  A frequent object of study in linguistic typology is the order of elements \{demonstrative, adjective, numeral, noun\} in the noun phrase. The goal is to predict the relative frequencies of these orders across languages. Here we use Poisson regression to statistically compare some prominent accounts of this variation. We compare feature systems derived from \citet{cinque2005deriving} to feature systems given in \citet{cysouw2010dealing} and Dryer (in prep). In this setting, we do not find clear reasons to prefer the model of \citet{cinque2005deriving} or Dryer (in prep), but we find both of these models have substantially better fit to the typological data than the model from \citet{cysouw2010dealing}.
\end{abstract}

\section{Introduction}

A frequent object of study in linguistic typology is the variation in the order of elements inside the noun phrase (NP) across languages.
In particular, much work has focused on predicting the relative frequencies across languages of orders of the elements \{demonstrative, adjective, numeral, noun\}.
Table~\ref{tab:order-freqs} shows the relative frequencies of different orders for these elements across languages (assuming each language exhibits only one dominant order) according to data given in Dryer (in prep).
In this table, $D$ stands for demonstrative, $N$ stands for numeral, $A$ stands for adjective, and $n$ stands for noun.
Genera counts are the counts of linguistic genera showing a certain
order; adjusted frequencies are calculated using a methodology
described in Dryer (in prep) intended to minimize any
overrepresentation of some orders that may arise from areal effects.

\begin{table}
  \centering
  \begin{tabular}{|l|r|r|}
    \hline
    Order & Adjusted frequency & Genera count \\
    \hline
    nAND &               43.50 &    84 \\
    DNAn &               36.62 &    57 \\
    DnAN &               28.34 &    38 \\
    DNnA &               21.18 &    31 \\
    NnAD &               15.33 &    28 \\
    nADN &               14.78 &    19 \\
    nDAN &                9.00 &    11 \\
    nNAD &                9.00 &     9 \\
    DnNA &                8.77 &    10 \\
    DAnN &                6.11 &     8 \\
    nDNA &                4.67 &     5 \\
    NAnD &                4.00 &     5 \\
    AnND &                3.00 &     3 \\
    NnDA &                3.00 &     3 \\
    NDAn &                3.00 &     3 \\
    AnDN &                2.49 &     3 \\
    DANn &                2.00 &     2 \\
    nNDA &                1.00 &     1 \\
    NADn &                0.00 &     0 \\
    NDnA &                0.00 &     0 \\
    ADnN &                0.00 &     0 \\
    ADNn &                0.00 &     0 \\
    ANDn &                0.00 &     0 \\
    ANnD &                0.00 &     0 \\
    \hline
  \end{tabular}
  \caption{NP orders and their adjusted frequencies across languages and their counts across genera, as given in Dryer (in prep).}
  \label{tab:order-freqs}
\end{table}

Here we consider three proposals from the literature on how to explain these frequencies.
We compare these proposals statistically in a log-linear framework based on how well they can predict the typological data given in Table~\ref{tab:order-freqs}.

We consider three proposals from the literature: those given in Dryer (in prep), \citet{cysouw2010dealing} and \citet{cinque2005deriving}.
The proposal in Dryer (in prep) is an update from the previous proposal of \citet{dryer2006cinque}.
The first two of these theories are featural in nature: they associate each order with a set of marked features, and claim that orders with more marked features will be less frequent.
The last model, that of \citet{cinque2005deriving}, is derivational in nature: it gives a generative model for how certain word orders arise, where certain decisions in the generative process are considered marked. Orders that require more marked operations to be generated are claimed to be less frequent.
We reduce the last model to a featural model, and then compare which model provides a feature system which can best predict the typological data when the features have different degrees of markedness.

\section{Method}

\subsection{Basics}
We consider each proposal from the literature to define a feature system, and compare the ability of each feature system to predict the observed frequencies of orders.
To do so, we use Poisson regression, as first used in \citet{cysouw2010dealing}.
In Poisson regression we represent each language with a set of $m$ binary-valued features, and say that the expected frequency $\lambda$ of a language in a sample of $k$ languages is given by:
\begin{align}
  \nonumber
  &\lambda = e^V \\
  \label{eq:V}
  &V = w_b + w_1 \cdot f_1 + w_2 \cdot f_2 + ... + w_m \cdot f_m,\\
\intertext{
  where $f_i$ is an indicator variable with value $0$ when the $i$th feature is $-$, and $1$ when the $i$th feature is $+$, and where the weights $w_b$ and $w_1, ..., w_m$ are those that maximize the probability of the observed counts of languages.
  The weight $w_b$ is called a \textbf{bias} term.
Under the probabilistic model of Poisson regression, the probabability that 
a language with feature values $f_1, ..., f_m$ has frequency $F$ is:}
  \label{eq:p}
  &p(F|f_1, ..., f_m) = \frac{\lambda^F e^{-\lambda}}{F!}.
\end{align}

Fitting a Poisson regression model using a certain set of features to
a set of (possibly adjusted) frequency tells us how well it is
possible to predict languages in this framework given that set of
features.  Feature weights may be negative, in which case they can be
considered \emph{marked}. In that case, the model embodies the claim
that the presence of these features is disfavored in languages.  Since
features get different weights, the model implements different degrees
of markedness per feature, as in Harmonic Grammar
\citep{smolensky2006harmonic}.  The model finds the degrees of
markedness which best predict the data given the feature system.

\subsection{Feature systems under comparison}

Within this framework, we compare three feature systems: (1) the system in Dryer (in prep), (2) the system in \citet{cysouw2010dealing}, and (3) the theory of \citet{cinque2005deriving}. Cinque's theory is not phrased in terms of features, so we use two reductions of his theory to features: those presented in \citet{merlo2015predicting} and our own, shown in Figure~\ref{fig:cinque-model}. Our featurization of Cinque's theory closely parallels the featurization given in \citet{cysouw2010dealing}.

\subsection{Dependent variables}

We apply Poisson regression to predict two quantities. First, we try
to predict the \emph{adjusted frequency} of each order, as given in
Dryer (in prep) and shown in Table~\ref{tab:order-freqs}.  (We round
the adjusted frequencies to the nearest integer in order to satisfy
the Poisson regression assumption that the dependent variable is a
natural number.)  Second, we try to predict the counts of genera given
in the same paper.

\subsection{Basis for model comparison}

We compare models using log likelihood, the log probability assigned to the observed frequencies under the model.
A model fits the data well when it assigns high probability to the data, so high log likelihood indicates a good fit.
When log likelihood for different models is close, we can also compare them by their degrees of freedom, which is the number of free parameters in the model.
In general, simpler models with fewer parameters are preferable over ones with more parameters.

\subsection{Notes on Featurization of \citet{cinque2005deriving}}
\label{sec:notes-feat-citetc}

Special care is needed when reducing the theory of \citet{cinque2005deriving} to features so that it can be compared with the other theories in a regression framework.

\begin{figure}[t]
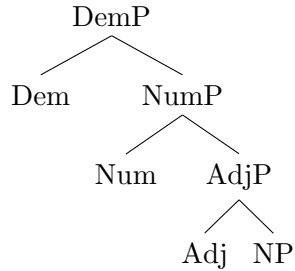

  \centering
  
\Tree [.DemP Dem [.NumP Num [.AdjP Adj NP ] ] ]

  \caption{The universal base structure of D,N,A,n under the theory of
    \citet{cinque2005deriving}.}
  \label{fig:cinque-base-order}
\end{figure}

The theory of \citet{cinque2005deriving} is not featural in nature,
but rather derivational. In this model, orders are built up by a
generative process that makes decisions in a certain order; whereas in
featural models, orders are assigned scores based on features that
have no intrinsic order.  For example, the centerpiece of the theory
of \citet{cinque2005deriving} is the claim that the Merge order of
D$>$N$>$A$>$n is universal, and that the Linear Correspondence Axiom
(LCA) of \citet{kayne1994antisymmetry} holds, such
that every word order must be generated from a base structure of the
form seen in Figure~\ref{fig:cinque-base-order}, plus movement
operations.  Orders that are not derivable from this base structure
are not generated at all in Cinque's theory.  For such orders, the
question of what if any movement operations apply never even arises
under Cinque's theory, in principle.

For example, suppose we think of the Cinque model in terms of features: then \alternativeMergeOrder is a feature that can be $+$ or $-$, and some movement operation is reflected in a feature that can be $+$ or $-$.
For an order that violates the specified merge order, we would give it value $+$ for \alternativeMergeOrder and $-$ for the movement feature, but this is not a completely correct reflection of Cinque's derivational theory.
The reason is that under the generative process, if a word order
violates the required merge order, then the model never even decides whether to perform
a movement operation or not: thus the value of the movement feature
should not be $-$ or $+$, but undefined.  

The theory of \citet{cinque2005deriving} also involves theoretically-derived graded markedness values for operations in the generative process: for instance, total movement is claimed to be unmarked, ``picture of who'' type movement is claimed to be especially marked, and the rest of the movement features are claimed to be marked.
In our methodology, we let the model decide on feature weights (markedness values) without regard to these a priori markedness values.
As such, it is possible that our implementation of the Cinque model
does not reflect its full intent. 

The fact that our weights are derived through fits to the data and not through a priori considerations is especially noteworthy in the case of the feature \alternativeMergeOrder in Cinque's system.
In a literal reading of \citet{cinque2005deriving}, orders which violate the required merge order should occur with probability 0, and thus the feature \alternativeMergeOrder should be assigned infinitely negative weight.
In that case the model would assign probability 0 to any data that has nonzero frequency for any such order.
In our work we let the model learn that \alternativeMergeOrder has a large negative weight, without postulating that orders violating the required merge order must have probability 0.

Here we represent Cinque's model using features for the sake of
convenience in statistical comparison, while noting that this
introduces the issues above.  And there is a further issue that should
be noted.  If Merge orders other than that seen in
Figure~\ref{fig:cinque-base-order} are allowed, then not only are
otherwise impossible word orders generable by postulating that those
word orders are generated as a base structure; additionally, word
orders that were already generable under Cinque's theory also wind up
with new derivations from different base structures.  Taking this
multiplicity of possible derivations into account increases the
complexity of the problem of inferring feature weights, and takes it
outside the scope of standard Poisson regression or other generalized
linear models.  This is a difficulty shared by the modeling approaches
of \citet{cysouw2010dealing} and \citet{merlo2015predicting}.  For
expediency, however, we follow previous work in treating every word
order not derivable from the D$>$N$>$A$>$n base structure of
Figure~\ref{fig:cinque-base-order} as being derived by an alternative
Merge order (so that \alternativeMergeOrder\texttt{=T}) with no
movement, ignoring alternative derivations.

We also note that in formalizing Cinque's (2005) theory into a
featural description, we noticed certain unclarities in the text which
affected our formalization.  These are as follows (note that Cinque
uses \textbf{N} where we use \textbf{n}, and \textbf{Num} where we use
\textbf{N}):
\begin{itemize}
\item Cinque describes order AnDN (his (k)) as involving two marked
  options: ``derivation with raising of NP plus \textbf{pied-piping of
    the picture of who type} of the lowest modifier (A), followed by
  \textbf{raising} of [A N] \textbf{without pied-piping} around both
  Num and Dem'' (emphasis ours).  Technically speaking, it is not
  clear whether this latter raising (without pied-piping) should count
  as marked in his system, since the only relevant parameter of
  movement is ``Movement of NP without pied piping'' (his (7biii)),
  but this latter raising is movement of AP, not NP.  Nevertheless we
  followed Cinque in assigning this word order a $+$ value for
  movement (of NP) without pied piping.  This decision is also
  justified by Cinque's comment (p. 320) that his system is ``[k]eeping to the
  idea that\ldots postnominal orders are only a function of the
  raising of the NP (or of an XP containing the NP)\ldots''.

Additionally, there is an
   inconsistency in \citet{cinque2005deriving} between (7bv), where
   this order is stated to involve partial movement, and (6k), where
   partial movement is not listed as a type of markedness for this
   order. Here we went with (7bv) and listed this order as involving
   \verb-partial_move=+-; we believe that this treatment is the most
   globally consistent overall, on analogy with orders such as NnAD
   which Cinque treats as involving partial movement because there are
   multiple types of movement and the first (raising of NP around A)
   is only partial.
 \item As with AnDN, there is inconsistency between (7bv) and the
   word-order-specific description of AnND (6w): in the  former, this order is
   stated to involve partial movement of NP, but in the latter,
   partial movement is not mentioned as a type of markedness.  As with AnDN,
   we listed this order as involving \verb-partial_move=+-.   
\end{itemize}
Regarding the first two cases, it should be emphasized that
\citet{cinque2005deriving} is far from totally clear about what does and
does not count as partial (and thus marked) movement.  For example,
NnAD is described as involving partial (and thus marked) raising of NP
around A, followed by a second raising that gets the raised
constituent all the way to the left edge. But although nNAD likewise starts
with a partial raising of NP around A (and N) followed by a second
raising that gets the raised constituent all the way to the left edge,
it is not considered to involve partial movement.

Additionally, there is one case of what we believe is a coding error
by \citet{cysouw2010dealing} in his implementation of Cinque's model
(see his Appendix on page 284): 

\begin{itemize}
\item Cysouw encodes nNAD as involving NP movement with pied piping of
  the \emph{picture of who} variety, but Cinque describes this order
  ((6t)) as involving \emph{whose picture} pied piping instead, which
  seems correct to us.
\end{itemize}
%
%

\subsection{Comparison with \citet{merlo2015predicting}}

\citet{merlo2015predicting} conducts a study with similar aims to ours and uses featurizations more or less the same as what we've discussed above.
She uses features to predict frequency classes using a Naive Bayes estimator and an Weighted Averaged One-Dependence (WAODE) estimator, rather than Poisson regression as we use here and as was proposed by \citet{cysouw2010dealing}.
As a summary of how this works: \citet{merlo2015predicting} first discretizes the integer-valued word order frequency counts (by language or by genus) into 2, 4, or 7 categories; then she learns a model that categorizes language classes by their features according to the classic Naive Bayes formula:
\begin{align*}
  \nonumber
  P(\text{class}|\text{features}) &\propto P(\text{features}|\text{class}) P(\text{class}) \\
  \nonumber
  P(\text{features}|\text{class}) &= \prod_{f_i \in \text{features}} P(f_i | \text{class}),
\end{align*}
where $f_i$ is the value of the $i$th feature in the featurization scheme under consideration; our $f_i$ here are Merlo's $a_i$ in her Equations 1--4, p. 334. A word order is assigned to the frequency class that maximizes the probability $P(\text{class}|\text{features})$ above for that word order's features.
Technically these ``features'' are attribute-value pairs, such as $\texttt{Symmetry1=T}$ for the Dryer model or $\texttt{Partial=whose-pp}$ for the Cinque model.
The WAODE model is a bit more complex than the Naive Bayes model but is fundamentally similar.

These models are trained under two approaches: type-based, where training data consists of language types and their features and frequencies, and token-based, where the frequent language types are repeated multiple times in the training data and the language types with frequency zero appear with frequency zero.
In the token-based approach, the model is not penalized for miscategorizing language types with frequency None, because these do not appear in the training data.

Our work differs from Merlo's approach on two points:
\begin{itemize}
\item In predicting typological data, we use Poisson regression, which is a discriminative log-linear predictor, rather than Naive Bayes and WAODE, which are generative models.
\item Our models predict integer-valued typological counts, whereas the models in \citet{merlo2015predicting} predict unordered categorically-valued frequency classes.
\end{itemize}

We favor Poisson regression (and more generally log-linear models) over the Naive Bayes/WAODE approach because it allows us to predict more fine-grained typological data and to model the strong intuition that the effects of features on typological frequencies should be monotonic.

Merlo's (2015) models have the property that feature weights are not monotonic in their preferences for frequency classes.
For example, in a model where the goal is to classify each language into the categories (e.g.) \{Very Frequent, Frequent, Rare, None\}, there is nothing to prevent a feature from getting weights that favor Very Frequent and None while disfavoring Frequent and Rare.
Examples of this non-monotonicity in feature weights can be seen in Merlo's (2015) Table 11, our Table~\ref{tab:merlo-table}: the feature $\texttt{Harmony=Y}$ favors a language to be either Very Frequent or None, while favoring Frequent and Rare less.
The monotonicity in weights means that the weights from this framework cannot be considered markedness values, which either penalize an order (make it less frequent) or do not.
In addition to making the model weights less interpretable, this non-monotonicity means that the model has the flexibility to take advantage of artifacts of the discretization of word order frequencies into bins.

\begin{table}
  \centering
  \begin{tabular}{|l r|}
    \hline
    Probability & Value \\
    \hline
    P($\texttt{Harmony=Y}|$Very Frequent) & = 0.99 \\
    P($\texttt{Harmony=Y}|$Frequent) & = 0.16 \\
    P($\texttt{Harmony=Y}|$Rare) & = 0.51 \\
    P($\texttt{Harmony=Y}|$None) & = 0.55 \\    
    \hline
  \end{tabular}
  \caption{Table of feature weights (conditional probabilities under the Naive Bayes assumption) from \citet{merlo2015predicting}, Table 11.}
  \label{tab:merlo-table}
\end{table}

\subsection{Comparison with \citet{cysouw2010dealing}}
\label{sec:comp-with-citetcys}

As stated in Section~\ref{sec:notes-feat-citetc}, our approach here is
very similar to that of \citet{cysouw2010dealing}: we use the same
statistical model class and the same theory comparison
(Cinque/Cysouw/Dryer).  The differences are as follows:
\begin{itemize}
\item We use the more recent data of Dryer (in prep) rather than the
  earlier data of Dryer (2006);
\item We use the feature set of Dryer (in prep) rather than the
  earlier feature set of Dryer (in prep);
\item We correct what we believe is a featurization error made by
  Cysouw in featurizing Cinque's theory.
\end{itemize}

\section{Results}

The results do not give clear grounds for deciding between the Dryer model and the Cinque model, but both of these models come out better than the \citet{cysouw2010dealing} model. Whether or not the Dryer model comes out better depends on whether we use the model to predict adjusted frequencies or genera counts.

\subsection{Predicting Adjusted Frequencies}

Table~\ref{tab:af-likelihoods} shows log likelihoods for models predicting adjusted frequency (rounded to the nearest integer). It also shows the number of parameters (d.f.) in each model. The table shows that Cinque's model slightly outperforms Dryer (in prep) on fitting the data.

\begin{table}
  \centering
  \begin{tabular}{|l|r|r|}
    \hline
    Model & Log likelihood & d.f. \\
    \hline
    Dryer (in prep) & -54.3 & 6 \\
    \citet{cysouw2010dealing} & -77.2 & 5 \\
    \citet{cinque2005deriving} (our features) & -53.0 & 8 \\
    \citet{cinque2005deriving} (Merlo's features) & -56.5 & 8\\
    \hline
  \end{tabular}
  \caption{Log likelihoods of \emph{adjusted frequency} data under various models, and the degrees of freedom (d.f.) of those models.}
  \label{tab:af-likelihoods}
\end{table}

\begin{figure}[ht!]
  \centering
  \includegraphics[scale=.7]{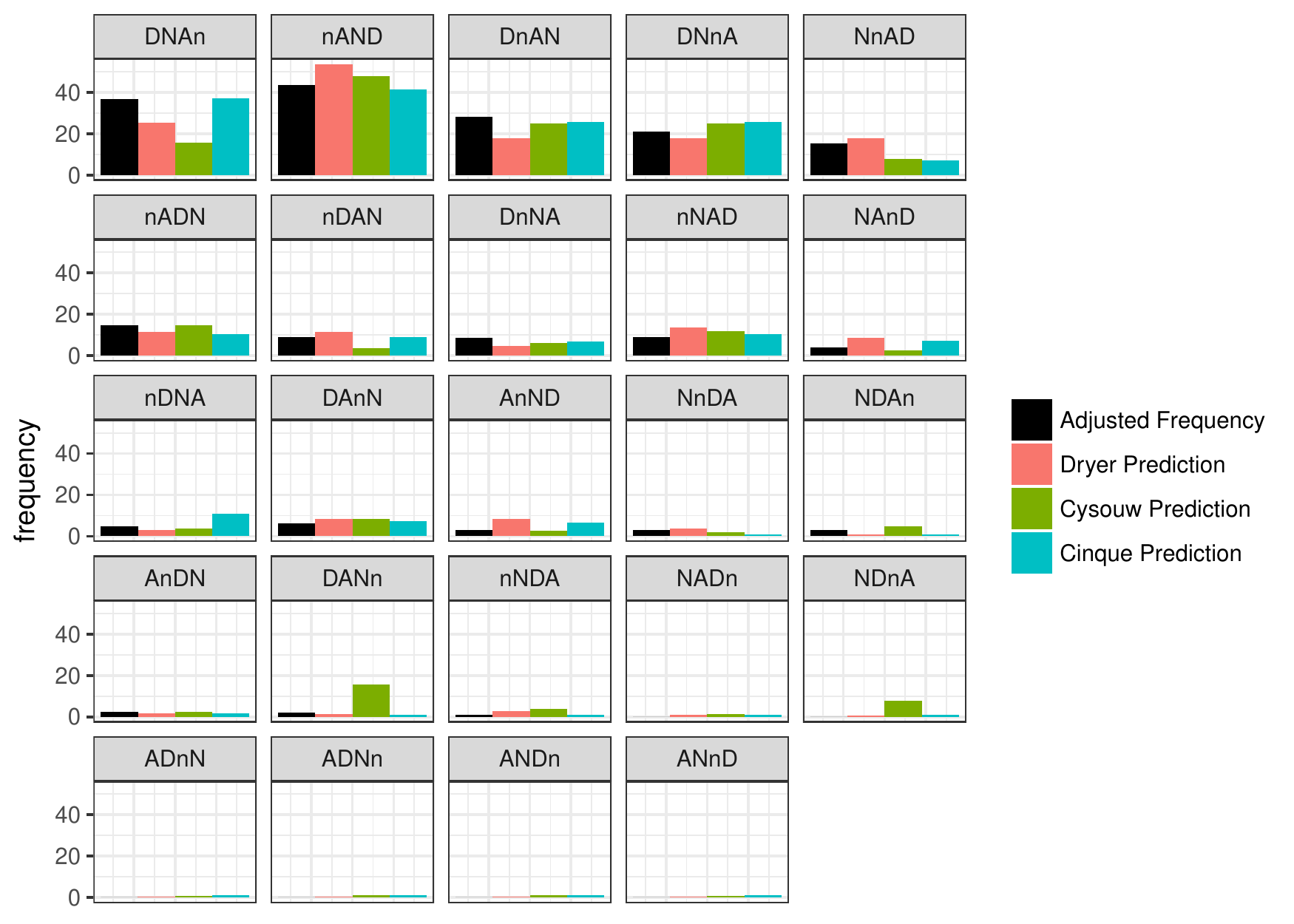}
  \caption{Adjusted frequency of word orders compared to model predictions. In this and all plots, the ``Cinque'' model refers to our featurization of Cinque's theory, including all features in the model without tying weights.}
  \label{fig:af-predictions}
\end{figure}

For a more detailed comparison of model performance, we compared model predictions to observed adjusted frequencies from Dryer (in prep). Figure~\ref{fig:af-predictions} shows model predictions compared against adjusted frequency.

\begin{figure}[ht!]
  \centering
  \includegraphics[scale=.7]{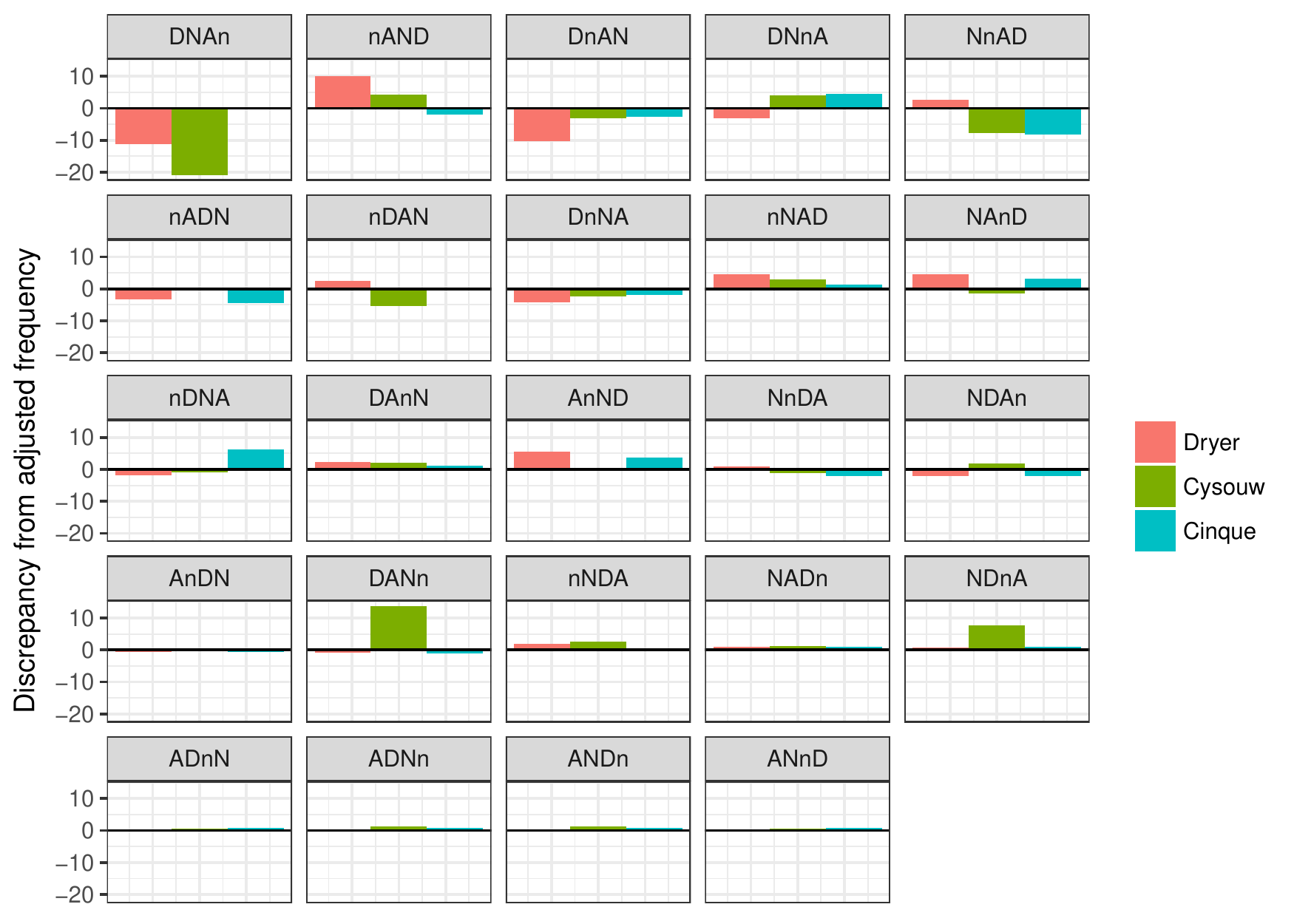}
  \caption{Signed chi-squared discrepancies between model predictions and adjusted frequency. Scores below zero mean that a model underpredicts frequency; scores above zero means that the model overpredicts frequency.}
  \label{fig:af-discrepancies}
\end{figure}

We wanted to know how much each order contributed to model fit, so in Figure~\ref{fig:af-discrepancies} we show signed $\chi^2$-discrepancies between model predictions and adjusted frequency. The $\chi^2$ discrepancy measures how much the prediction error for each word order contributes to the overall discrepancy between data and model fit; signed $\chi^2$ discrepancy presents this discrepancy in the direction of the discrepancy for each order (whether it under-predicts or over-predicts).
If a model predicts a count of $E_i$ for the $i$th word order and the observed count is $O_i$, then the signed $\chi^2$ discrepancy is:
\begin{align}
  \nonumber
  \frac{(O_i - E_i) \times |O_i - E_i|}
       {E_i}.
\end{align}
The magnitude of the discrepancy corresponds to how much a model is penalized for failing to predict a certain order.

\begin{figure}[ht!]
  \centering
  \includegraphics[scale=.7]{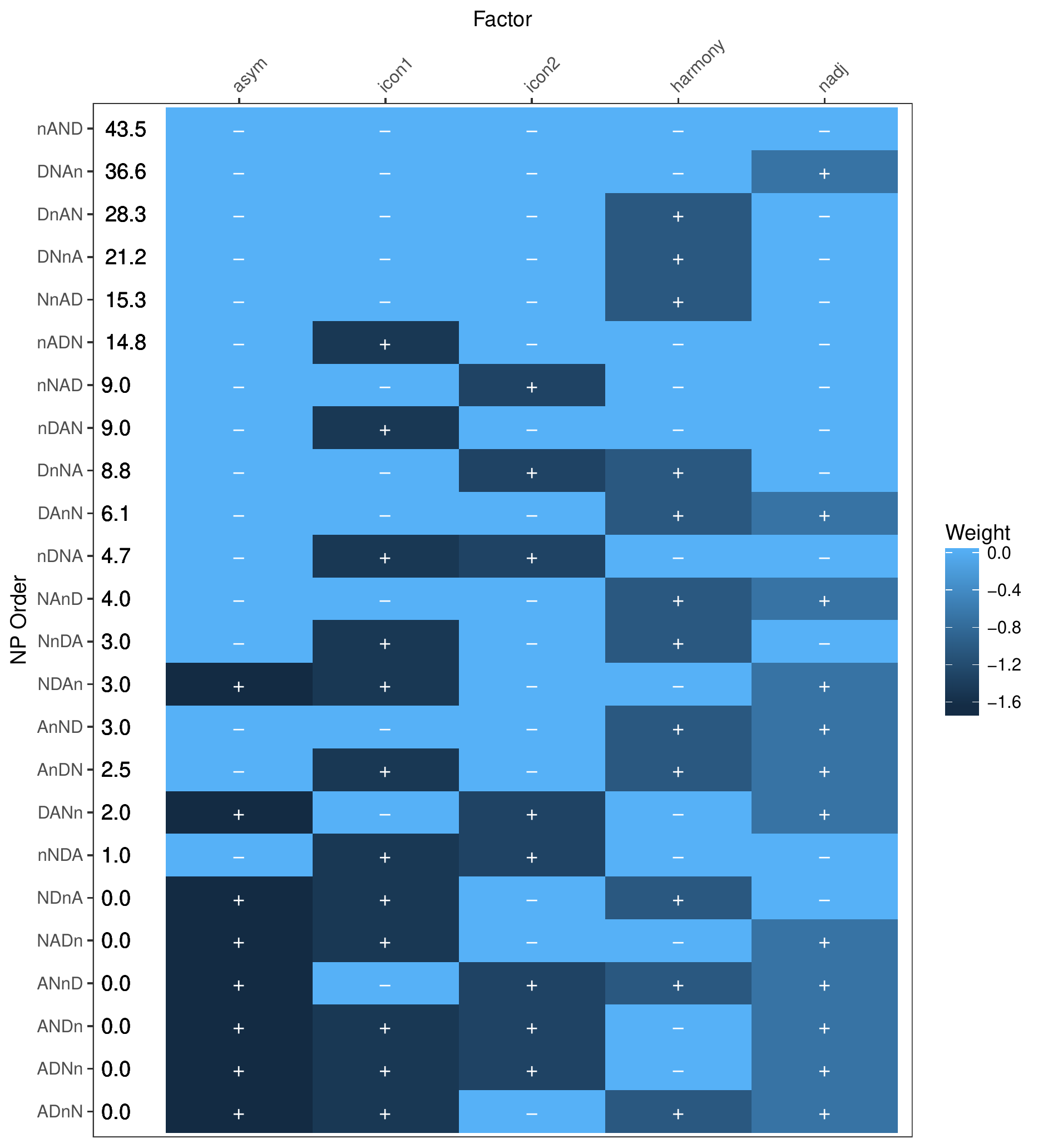}
  \caption{Feature weights from the Dryer (in prep) model when predicting adjusted frequency (first column).}
  \label{fig:dryer-model}
\end{figure}

\begin{figure}[ht!]
  \centering
  \includegraphics[scale=.7]{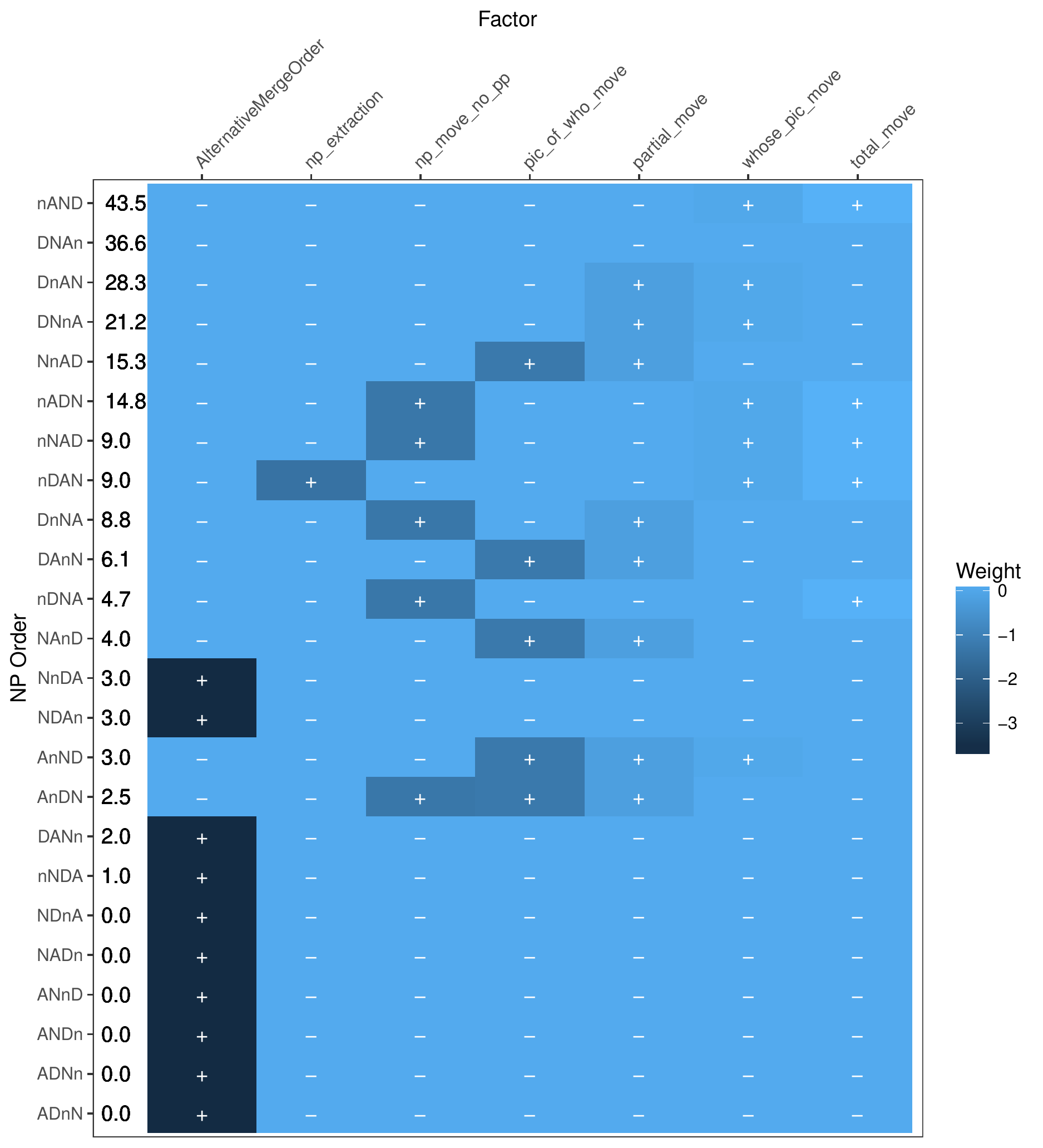}
  \cprotect\caption{Feature weights from the \citet{cinque2005deriving} model when predicting adjusted frequency (first column).}
  \label{fig:cinque-model}
\end{figure}

As another way to analyze the results, we show in Figures~\ref{fig:dryer-model} and \ref{fig:cinque-model} the weights assigned to features for different word unders under the Dryer model and the Cinque model.
Here we see that Cinque's model works by strongly penalizing low-frequency orders using the \alternativeMergeOrder feature, and then the differences among the remaining orders are handled by the rest of the features.

One limitation of applying Poisson regression to the adjusted
frequency data is that Dryer (in prep)'s method of computing adjusted
frequencies in general compresses high frequency counts more than low
frequency counts.  This means that adjusted frequency may overly penalize
models that perform best at predicting the counts of common orders.
For this reason, it is also important to evaluate the performance of the models under
consideration in predicting genera counts.  We turn to this matter in
the next section.

\subsection{Predicting Genera Counts}

Now we turn to models that were trained to predict the genera count data given in Dryer (in prep).
When we use the various feature systems to predict genera counts, we get the following data log-likelihoods, shown in Table~\ref{tab:genera-likelihoods}:

\begin{table}
  \centering
  \begin{tabular}{|l|r|r|}
    \hline
    Model & Log likelihood & d.f. \\
    \hline
    Dryer (in prep) & -61.3 & 6 \\
    \citet{cysouw2010dealing} & -106.0 & 5 \\
    Cinque (our features) & -67.7 & 8 \\
    Cinque (Merlo's features) & -71.1 & 8 \\
    \hline 
  \end{tabular}
  \caption{Log likelihoods of \emph{genera count} data under various models, and the degrees of freedom (d.f.) of those models.}  
  \label{tab:genera-likelihoods}
\end{table}

So when predicting genera, we get the best fit to the data using the set of features from Dryer (in prep), followed by Cinque's (2005) features, followed by Cysouw's (2010) features.

\begin{figure}[ht!]
  \centering
  \includegraphics[scale=.7]{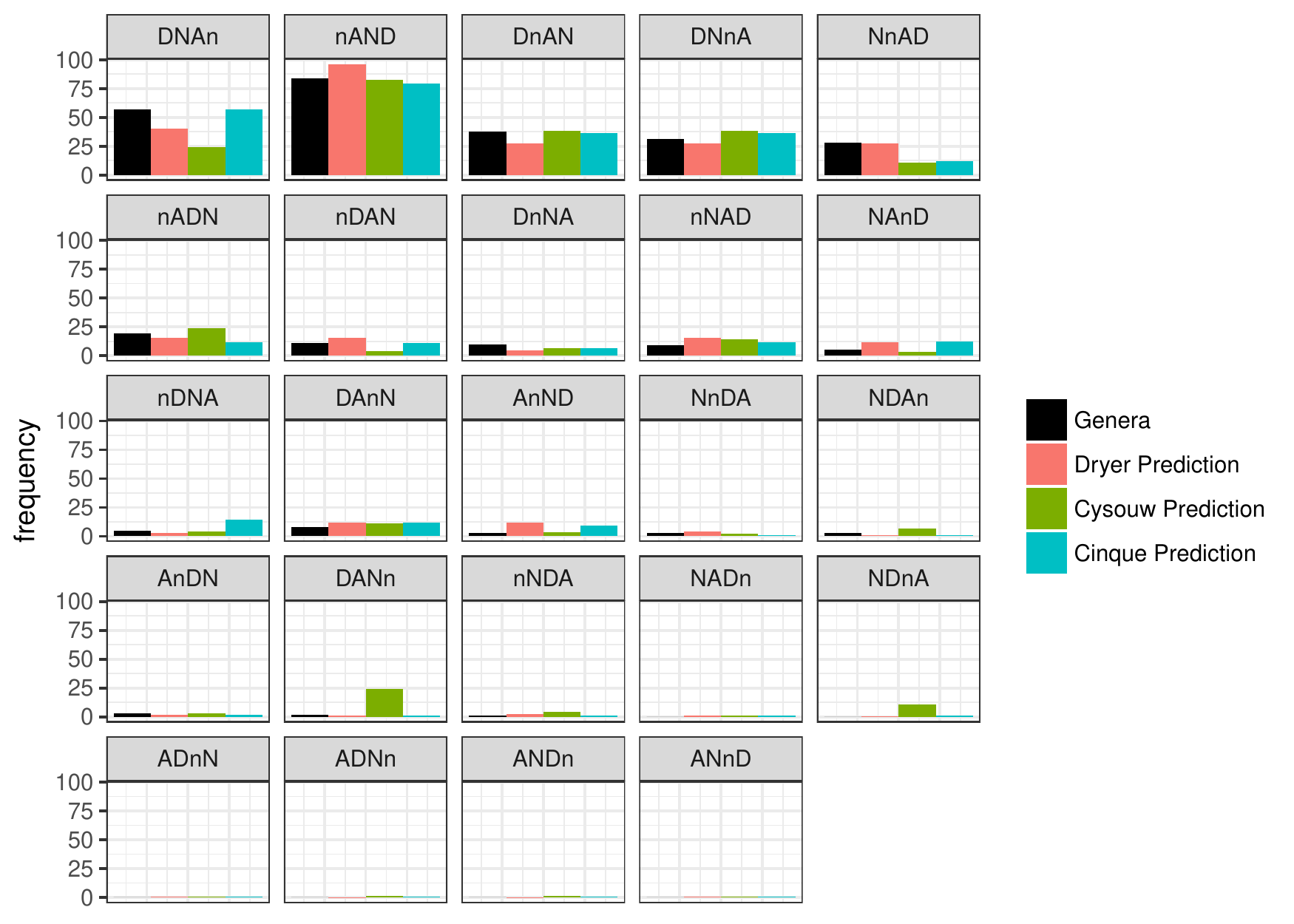}
  \caption{Genera counts of word orders compared to model predictions. In this and all plots, the ``Cinque'' model refers to our featurization of Cinque's theory, including all features in the model without tying weights.}
  \label{fig:genera-predictions}
\end{figure}

\begin{figure}[ht!]
  \centering
  \includegraphics[scale=.7]{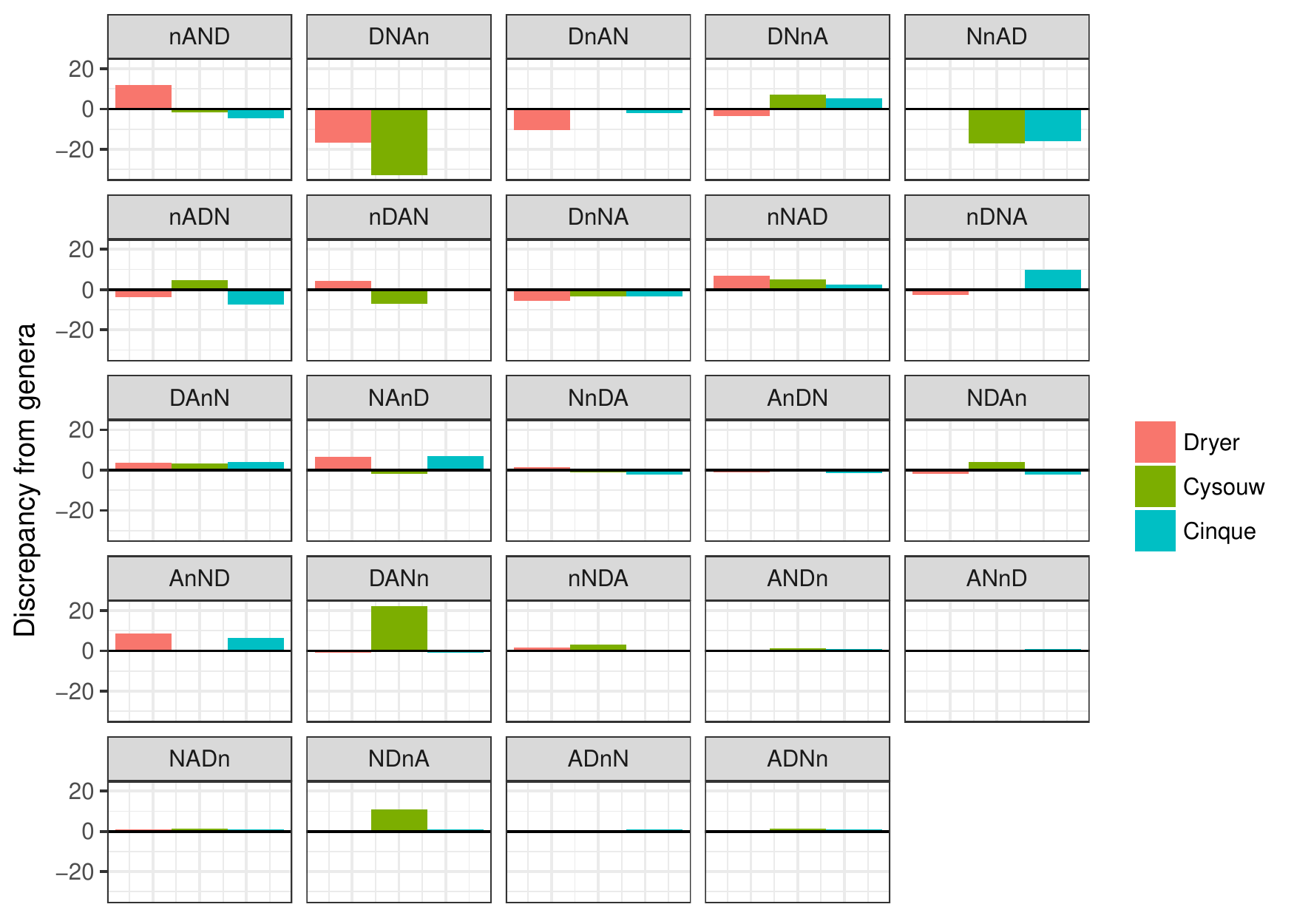}
  \caption{Signed chi-squared discrepancies between model predictions and genera counts. Scores below zero mean that a model underpredicts frequency; scores above zero means that the model overpredicts frequency.}
  \label{fig:genera-discrepancies}
\end{figure}

\begin{figure}[ht!]
  \centering
  \includegraphics[scale=.7]{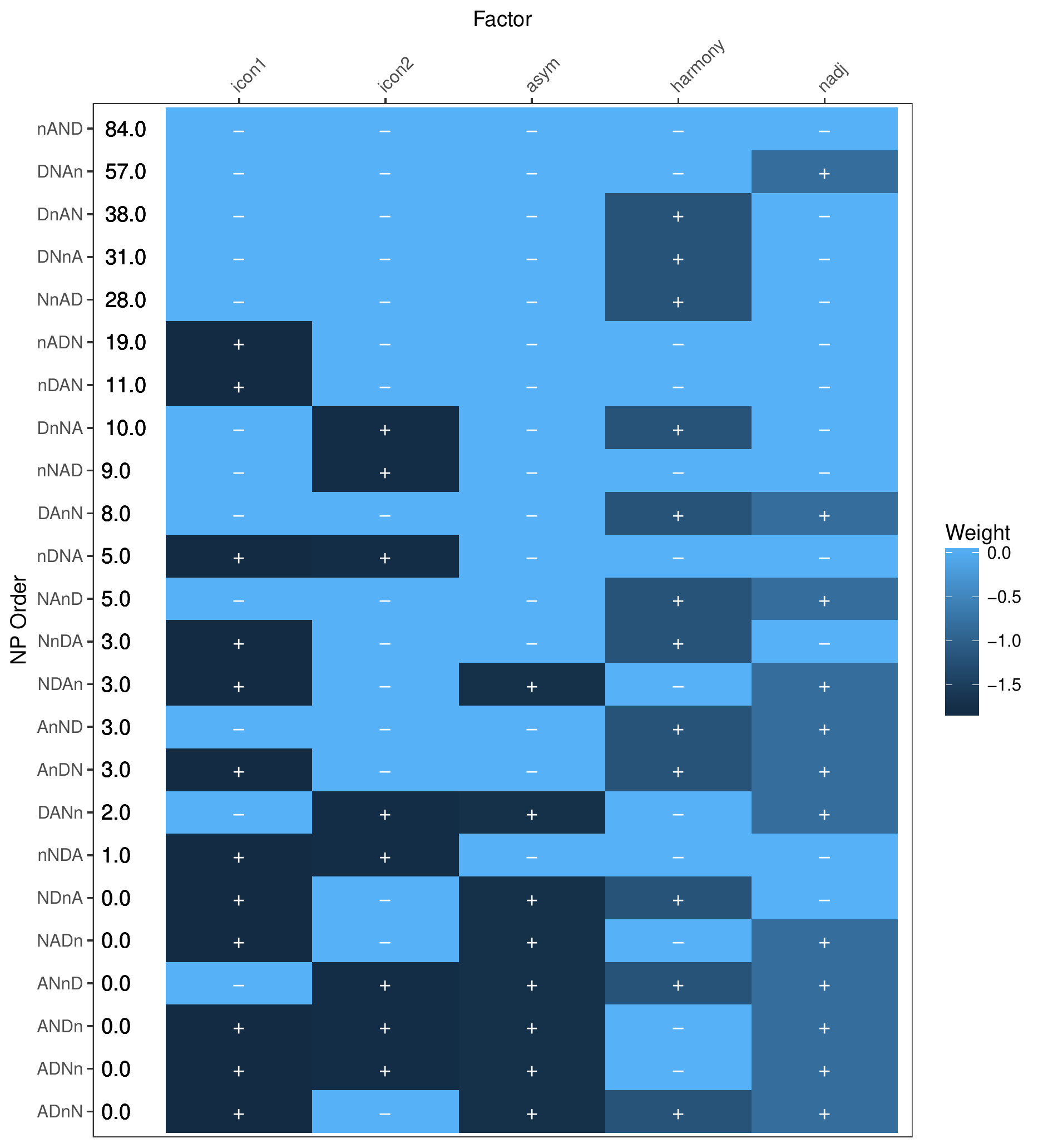}
  \caption{Feature weights from the Dryer (in prep) model when predicting genera counts (first column).}
  \label{fig:dryer-model-genera}
\end{figure}

\begin{figure}[ht!]
  \centering
  \includegraphics[scale=.7]{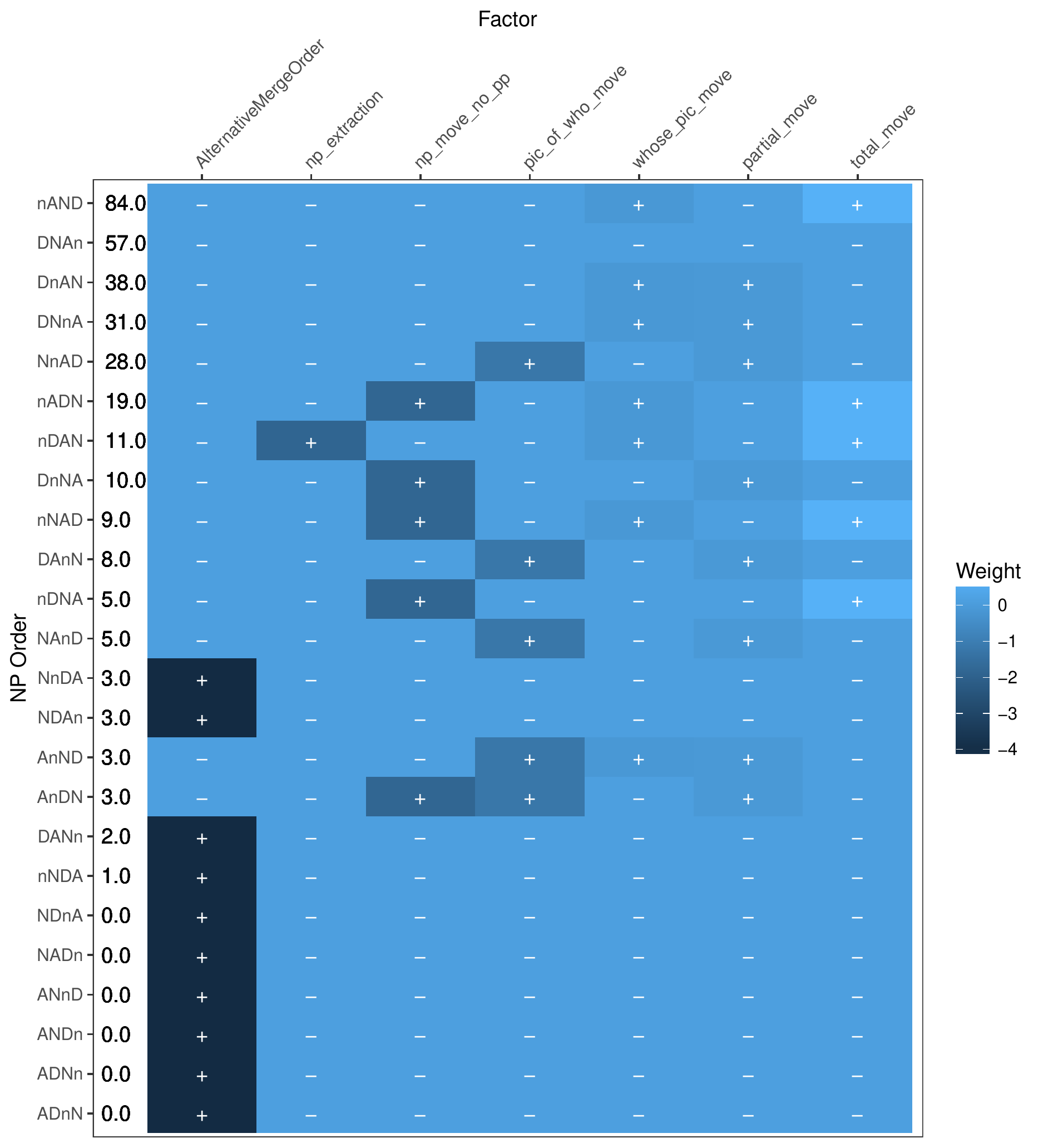}
  \cprotect\caption{Feature weights from the \citet{cinque2005deriving} model when predicting genera counts (first column). The feature \verb+total_move+ comes out to have a (non-significant) positive weight.}
  \label{fig:cinque-model-genera}
\end{figure}

We think Cinque's model comes out worse when predicting genera primarily because it underpredicts $NnAD$ orders, whereas the Dryer model gets that order exactly correct. This can be seen in Figure~\ref{fig:genera-predictions}, which shows model predictions, and Figure~\ref{fig:genera-discrepancies}, which shows signed $\chi^2$ discrepancies compared to genera counts.
Figures~\ref{fig:dryer-model-genera} and \ref{fig:cinque-model-genera} show the optimal feature weights for the Dryer and Cinque models, respectively, when predicting genera counts.

\section{Discussion}

The results give clear evidence that the Dryer (in prep) and \citet{cinque2005deriving} model provide feature systems that have better predictive power than the model of \citet{cysouw2010dealing}. But in our opinion they do not give strong reason to favor Cinque's model over Dryer's model or vice versa. Although under a certain interpretation Cinque's model can provide a slightly higher fit to the data, this only holds under one featurization, and it does not hold when predicting genera counts. The discrepancy in results between adjusted frequency and genera counts may be due to the particular distributional characteristics of adjusted frequency as discussed above. The analysis suggests overall that the Dryer model and the Cinque model have roughly similar predictive power, and the current data do not discriminate between them.

\section*{Acknowledgments} 

This work was supported by NSF DDRI grant \#1551543 to R.F.

\nocite{dryer2017order}

\bibliography{writeup}
\bibliographystyle{apa}

\appendix
\section{Feature weights}
\begin{table}[ht!]
  \centering
  \begin{tabular}{|l|r|r|r|}
    \hline
    Feature & Weight & Std. Error & $p$ \\
    \hline
    (Bias) &  3.9815  &   0.1175 &  $\lt$ .001 \\
    icon1  & -1.5382  &   0.1910 &  $\lt$ .001 \\
    icon2  & -1.3726 &      0.2177 &  $\lt$ .001 \\
    asym  &  -1.7200 &      0.4846 &  $\lt$ .001 \\
    harmony &     -1.0936   &  0.1436 & $\lt$ .001 \\
    nadj    &     -0.7480  &   0.1641 & .005    \\
    \hline
  \end{tabular}
  \caption{Feature weights for the model of Dryer (in prep) when predicting \textbf{adjusted frequency}.}
\end{table}

\begin{table}[ht!]
  \centering
  \begin{tabular}{|l|r|r|r|}
    \hline
    Feature & Weight & Std. Error & $p$ \\
    \hline
    (Bias)      &  3.8649 &     0.1165 &  $\lt$ .001 \\
    na\_adjacency &  -1.3835 &     0.1917 & $\lt$ .001 \\
    n\_edge      &  -0.6402 &     0.1415 & $\lt$ .001 \\
    d\_edge      &  -1.1876 &     0.1561 & $\lt$ .001 \\
    na\_order    &  -1.1096 &     0.1584 & $\lt$ .001 \\
    \hline
  \end{tabular}
  \caption{Feature weights for the model of \citet{cysouw2010dealing} when predicting \textbf{adjusted frequency}.}
\end{table}

\begin{table}[ht!]
  \centering
  \begin{tabular}{|l|r|r|r|}
    \hline
    Feature & Weight & Std. Error & $p$ \\
    \hline
    (Bias)              &  3.61092 &    0.16440 &  $\lt$ .001 \\
    AlternativeMergeOrder & -3.71628  &  0.37167 & $\lt$ .001 \\
    whose\_pic\_move       &  -0.06587 &  0.26465 & .803 \\
    np\_move\_no\_pp       &  -1.39843 &  0.23706 & $\lt$ .001 \\
    pic\_of\_who\_move     &  -1.33916 &  0.28958 & $\lt$ .001 \\
    partial\_move        &  -0.30165 &    0.32682 & .356 \\
    np\_extraction       &  -1.52919 &    0.36381 & $\lt$ .001 \\
    total\_move          &  0.18137 &    0.35649 & .611 \\
    \hline
  \end{tabular}
  \caption{Feature weights for the model of \citet{cinque2005deriving} under our featurization when predicting \textbf{adjusted frequency}.}
\end{table}

\begin{table}[ht!]
  \centering
  \begin{tabular}{|l|r|r|r|}
    \hline
    Feature & Weight & Std. Error & $p$ \\
    \hline
    (Bias)                    &  3.39905 &    0.14848 &  $\lt$ .001 \\
    AlternativeMergeOrder & -3.51612 &   0.37858 & $\lt$ .001 \\
    partial\_np                     &  -1.35184 &    0.23373 & $\lt$ .001 \\
    partial\_of-who-pp              &  -2.05580 &    0.35159 & $\lt$ .001  \\
    partial\_whose-pp                &  -0.00645 &   0.17115 & .970 \\
    complete\_np                    &  -1.17525 &    0.20835 & $\lt$ .001 \\
    complete\_of-who-pp             &  -1.16650 &    0.24160 & $\lt$ .001 \\
    complete\_whose-pp               &  0.29933 &    0.19247 & .120 \\
    \hline
  \end{tabular}
  \caption{Feature weights for the model of \citet{cinque2005deriving} under Merlo's (2015) featurization when predicting \textbf{adjusted frequency}.}
\end{table}

\begin{table}[ht!]
  \centering
  \begin{tabular}{|l|r|r|r|}
    \hline
    Feature & Weight & Std. Error & $p$ \\
    \hline
    (Bias) & 4.56458 &    0.09102 &  $\lt$ .001 \\
    icon1      & -1.84600 &    0.17029 & $\lt$ .001 \\
    icon2      & -1.80865 &    0.20695 & $\lt$ .001 \\
    asym       & -1.77394 &    0.47645 & $\lt$ .001 \\
    harmony    & -1.25179 &    0.11740 & $\lt$ .001 \\
    nadj       & -0.86771 &    0.13439 & $\lt$ .001 \\
    \hline
  \end{tabular}
  \caption{Feature weights for the model of Dryer (in prep) when predicting \textbf{genera counts}.}
\end{table}

\begin{table}[ht!]
  \centering
  \begin{tabular}{|l|r|r|r|}
    \hline
    Feature & Weight & Std. Error & $p$ \\
    \hline
    (Bias)      & 4.41097 &    0.09217 &  $\lt$ .001 \\
    na\_adjacency & -1.77523 &    0.17740 &  $\lt$ .001 \\
    n\_edge       & -0.76628 &    0.11678 & $\lt$ .001 \\
    d\_edge       & -1.24504 &    0.13132 &  $\lt$ .001 \\
    na\_order     & -1.22424 &    0.13152 &  $\lt$ .001 \\
    \hline
  \end{tabular}
  \caption{Feature weights for the model of \citet{cysouw2010dealing} when predicting \textbf{genera counts}.}
\end{table}

\begin{table}[ht!]
  \centering
  \begin{tabular}{|l|r|r|r|}
    \hline
    Feature & Weight & Std. Error & $p$ \\
    \hline
    (Bias)                & 4.0431 &     0.1325 & $\lt$ .001\\
    AlternativeMergeOrder & -4.1484 &     0.3586 & $\lt$ .001\\
    whose\_pic\_move        & -0.2471 &     0.2335 & .290 \\
    np\_move\_no\_pp         & -1.9358 &     0.2081 & $\lt$ .001\\
    pic\_of\_who\_move       & -1.3523 &     0.2527 & $\lt$ .001\\
    np\_extraction         & -1.9767 &     0.3203 & $\lt$ .001\\
    partial\_move          & -0.2081 &     0.2835 &   .463 \\
    total\_move            &  0.5786 &    0.2956  &  .050 \\
    \hline
  \end{tabular}
  \caption{Feature weights for the model of \citet{cinque2005deriving} under our featurization when predicting \textbf{genera counts}.}
\end{table}

\begin{table}[ht!]
  \centering
  \begin{tabular}{|l|r|r|r|}
    \hline
    Feature & Weight & Std. Error & $p$ \\
    \hline
    AlternativeMergeOrder & -3.92246 &    0.37106 & $\lt$ .001 \\
    partial\_np                      & -1.79085 &    0.21596 & $\lt$ .001 \\
    partial\_of-who-pp               & -2.40130 &    0.31357 & $\lt$ .001 \\
    partial\_whose-pp                & -0.03155 &    0.14208 & .824 \\
    complete\_np                     & -1.35753 &    0.18655 & $\lt$ .001 \\
    complete\_of-who-pp              & -0.98344 &    0.18786 & $\lt$ .001 \\
    complete\_whose-pp                & 0.52716 &    0.15617 & $\lt$ .001 \\
    \hline
  \end{tabular}
  \caption{Feature weights for the model of \citet{cinque2005deriving} under Merlo's (2015) featurization when predicting \textbf{genera counts}.}
\end{table}

\end{document}